\documentclass[journal]{IEEEtran}
\usepackage{xstring}

\newcommand*{\ALL}{}%

\ifdefined\ALL
    \newcommand{\subsubsectionc}{\subsubsection}
        \newcommand{\subsectionc}{\subsection}
\else
    \newcommand{\subsubsectionc}[1]{}
        \newcommand{\subsectionc}[1]{}
\fi

\usepackage[utf8]{inputenc}
\usepackage{booktabs}
\usepackage{amsmath}    
\usepackage{url}
\usepackage{multirow}

\usepackage{graphicx}
\ifCLASSINFOpdf
\else
\fi
\begin{document}
%
\title{Human activity recognition from skeleton poses}
%
%
%

\author{Frederico Belmonte Klein,
Angelo Cangelosi,~\IEEEmembership{Senior Member,~IEEE}
\thanks{Frederico Belmonte Klein (e-mail: frederico.klein@plymouth.ac.uk), from School of Computing, Electronics and Mathematics, Plymouth University PL4 8AA, Plymouth, UK

}
\thanks{Angelo Cangelosi (e-mail: a.cangelosi@plymouth.ac.uk), from School of Computing, Electronics and Mathematics, Plymouth University PL4 8AA, Plymouth, UK}

}

%
%


\markboth{ Unpublished } {}

%



\maketitle

\begin{abstract}
Human Action Recognition is an important task of Human Robot Interaction as cooperation between robots and humans requires that artificial agents recognise complex cues from the environment. 
A promising approach is using trained classifiers to recognise human actions through sequences of skeleton poses extracted from images or RGB-D data from a sensor. 
However, with many different data-sets focused on slightly different sets of actions and different algorithms it is not clear which strategy produces highest accuracy for indoor activities performed in a home environment. 
This work discussed, tested and compared classic algorithms, namely, support vector machines and k-nearest neighbours, to 2 similar hierarchical neural gas approaches, the growing when required neural gas and the growing neural gas.  
\end{abstract}


%
\IEEEpeerreviewmaketitle

\section{Introduction}
%
%
%
%
\IEEEPARstart{A}{s} world population is ageing, social and health care of older adults becomes of increasing concern. 
One suggestion on how to support longer independent living is by using the socially assistive robots. 
In this context, we propose to implement a desirable feature for independent living: detect  activities  performed by older adults and provide some aid, if needed, by using a robot.

With this goal in mind, we have identified as a starting point the use of RGB-D sensor data and skeletons provided by the Microsoft KINECT sensor and chose the data-set CAD-60. 
We proceeded to implement one of the state of the art classifiers used to predict the human activities, namely the Growing When Required Neural Gas for activity detection, as it was implemented by Parisi et al. in 2015\cite{parisi}. We have replicated this approach and compared it to other simpler approaches such as K-Nearest Neighbours and Support Vector Machines.\footnote{The MATLAB code used for all the work done to classify the CAD-60 data-set is available for free in \url{https://github.com/frederico-klein/cad-gas}. }



 

Human Activity Recognition (HAR) is a very important intermediate goal, on which many tasks depend, if one is aiming to develop a capable and useful automatic assistant and/or service robot, specially in health-care applications. 
In the special case of care of older adults, it is fundamental that a helper system would work with as little as possible input form the user, in a virtually autonomous way, as one of the goals for such an assistant is that it would provide immediate and accurate assistance when the user may not be at their most wholesome state. 
In this sense, HAR plays a fundamental role and is desirable for it to be based on as little as possible actively gathered information. 
RGB-D (RGB plus Depth), by increasing visual information with depth sensing  provides a much simpler to analyse paradigm when compared with the challenge of normal 2D colour video. 
RGB-D data deserves special recognition for facilitating image segmentation and consequently making it easy to extract skeleton poses, which have been proven useful for HAR. Usage of RGB-D data from a KINECT sensor has been the initial paradigm \cite{parisi,chaquet_survey_2013, act_rgbd, gupta_human_2013,shan_3d_2014, faria_probabilistic_2014, cippitelli_human_2016} to implement activity recognition, most often utilizing skeleton sequences over time. 
Nowadays it is possible to accurately extract more detailed skeleton model information quite efficiently with convolutional network models such as OpenPose~\cite{gines_hidalgo_zhe_cao_tomas_simon_shih-en_wei_hanbyul_joo_yaser_sheikh_openpose:_????}, however, how exactly to put to use the knowledge about these poses is still an ongoing research topic. 

In this study, the extension of fall detection to multiple activities detection, as this is a more complex task, is planned to enable the testing of the fall detection, or better said, of the fall detection principle. 
One may achieve that by describing falling as a specific action and use other types of activities as proxies for a falling event. 
This step is perceived as necessary, as to get people to do authentic falls in laboratory environment is debatable. The alternative, that is, to test it in such an early stage a real home setting would be impractical given the rarity of falls. 
Even if a group of older adults with more than 65 years of age (which is epidemiologically more prone to falls) is chosen, we could only expect around 33\% chance of having a fall in a year~ \cite{berry_falls:_2008}. 
This figure lowers to around 15\% per year if a `healthy' population of older people is chosen~\cite{masud2001epidemiology}. 

Understandingly, testing fall detection indirectly with HAR may seem a bit counter-intuitive at first, but it has the advantage of allowing to test the components of the system separately. 
Deploying and testing a mature HAR system first has the advantage of enabling us to check: which HAR algorithm and strategy is the most accurate; which user's active tracking is the best; is it better to have a mobile robot or multiple sensors with pan-tilt units in key locations; what strategy is commercially viable and so on.  Only then with a mature system one could conceive to conduct long term and large population sized testing. 


Another major advantage of such approach is that, whilst humans would find very tedious the task of watching over a person for a whole year to identify the chance of fall, a device can examine every sample in the same manner. 
As a matter of fact, it is impractical, intrusive and not economically viable to use humans for surveillance, which makes this sort task only possible if achieved by automatic means. 
And while false positives are expected, checking through this smaller subgroup is expected to be less time-consuming, more accurate and more socially acceptable than being supervised continuously. 

Aside from social care assistance HAR may potentially have numerous medical uses. Foremost one may think of epidemiological studies, where this new data would be invaluable, opening  completely new fields of study which could correlate activity patterns and eating habits to illnesses. 
More specifically, one may expect detecting gait and movement alterations that require medical attention and may be first indicators of motor illnesses or neurodegenerative disorders, such as Parkinsonisms (where bradykinesia, rigidity and tremor could be identified and graded~\cite{gaenslen_early_2010,klucken_unbiased_2013}), or Alzheimer's (where we would maybe identify apraxia), or detect signs from syndromes such as gait apraxia or ataxia~\cite{cecil}. 
One may also track the severity of motor symptoms as well as correlate it to medication efficacy. 
It could help map activity patterns that may be first signs of dementia or depression, as well as be a part of a more complex telemedicine system, with early detection of heart problems~\cite{di_lenarda_future_2017}, strokes~\cite{demaerschalk_stroke_2009} or respiratory exacerbations for patients with COPD~\cite{cooper_respiratory_2009}. 
HAR could serve as early detector when symptoms of illness are subtle or a special aid in diagnosis when a certain disease is paroxysmal or occurs in a wax and wane patterns, where the health professional can examine the recorded data of an event. 
Although one must be cautious about the efficacy of telemedicine~\cite{flodgren_interactive_2015}, hopefully, with the development of better tools and larger studies, many of the foreseen advantages we expect from it can be achieved.

\section{Literature review}

A vast literature already exists on HAR through RGB-D data as it has in normal video (RGB) data~\cite{chaquet_survey_2013}. 
A high quality and thorough review was done by Zhang et. al. \cite{act_rgbd}, which relates the most widely used data-sets for activity detection as well as the benchmark holder algorithms for such data-sets. 
In this review it is documented that different activities have different levels of interaction between subjects, objects in the scene and the environment, as well as that different data-sets offer varying complexity with different number classes, which can be wider, full body, actions (kicking, running) or more finely grained actions (hand movements). 
We will focus on methods that use skeletons and try to classify whole body postures - mostly related to the initial goal task of classifying falls. 
Many other data-sets and algorithms could be tried, perhaps the most interesting one for our application would be the RGBD-SAR data-set, which presents older adults performing daily activities. 
However, this data-set does not provide ground truth skeletons and this would increase the difficulty, adding another layer of uncertainty in validating our classifier choices. 
We decided thus to use the CAD-60 for being widely studied and offering a larger range of previous work for frame of reference, as well as presenting actions that are relevant to care of older adults. 

As such, we will briefly report state of the art results of CAD-60 in chronological order. 
A paper along with the data-set was published in 2012 by Sung et al.~\cite{cad60} and it utilises a two-layered maximum entropy Markov model with a on-the-fly graph structure selection presenting a ``precision/recall of 84.7\%/83.2\% in detecting the correct activity when the person was seen before in the training set and 67.9\%/55.5\% when the person was not seen before''~\cite[p.~842]{cad60}. 
A comprehensive list of precision and recall values is available at Cornell's website~\cite{_cornell_????} for conference.

Many other classifiers for this data-set where published in the following years.  
In one of the first publications on this data-set, Gupta et al. in 2013~\cite{gupta_human_2013} classified this data-set without using the skeleton information, i. e. only the depth maps, using depth information for better segmentation and code descriptors to feed an ensemble discriminator achieving 78.1\% precision and 75.4\% recall. 
Most of the other listed classifiers and specially the ones with higher accuracy, used skeleton information for the classification task, to name a few:
Shan and Akella using skeleton information, in 2014~\cite{shan_3d_2014} implemented a classifier that estimates key poses based on estimation of kinetic energy and a support vector machine to achieve a global precision of 93.8\% and 94.5\% recall; 
Faria et al. in 2014~\cite{faria_probabilistic_2014} used a dynamic Bayesian network model to assign weights to multiple classifiers and implement an ensemble learning technique to achieve 91.1\% precision and 91.9\% recall overall.  

Of particular interest to us is the architecture that we were trying to replicate from Parisi et al. 2015~\cite{parisi}, which uses a chained growing when required neural gas classifier to achieve a global 91.9\% precision and 90.2\% recall on this data-set. 
After this publication, two very similar works deserve special note, namely the work of Cippitelli et al. in 2016~\cite{cippitelli_human_2016}, which uses a multi-class support vector machine with a radial basis function kernel to achieve a global 93.9\% precision and 93.5\% recall on the CAD-60 data-set. 
And more recently Manzi et al. in 2017~\cite{alessandro_manzi_human_2017} used a very similar approach, changing the k-means clustering to a x-means algorithm and a sequential minimal optimization process to train the SVM faster, achieved 100\% precision and recall on the CAD-60 data-set and a  93.3\% accuracy on the TSTv2 data-set. 
Note that our own work~\cite{klein_2016} on that data-set with the hierarchical GWR neural gas had an accuracy of 90.2\%.


Unlike in other areas of artificial intelligence and machine learning where algorithms are plentiful, freely available source code for HAR algorithms are hard to find, making it difficult to develop applications that depend on it. 
In this work are presented alternatives that are simple to implement and have reasonable performance, along with the source code. We also present some of the issues encountered, which should contribute to further development of this field.

\section{Methods}


For this work we chose to classify the activities from the CAD-60 (available at \cite{cad60}), which is
a data-set containing 4 subjects performing 12 different actions, captured by a KINECT version 1. 
The data-set contains 320x240 pixels RGB-D motion sequences and skeletal information acquired at a constant frame rate of 30fps~\cite{noauthor_KINECT_nodate}. 
The skeletal information is composed of 15 points per skeleton with x,y and z coordinates corresponding to the best estimate location of joint positions in a 3D space as extracted by its algorithm from the depth data. 
For the present work the RGB-D will be disregarded and only the skeleton joint positions will be used. 
 
To classify this data, 4 different algorithms were used, support vector machines (SVM), k-nearest neighbours (KNN) as well as hierarchical neural gas-based machines: the Growing Neural Gas (GNG) and the Growing When Required neural gas (GWR).  
Either matched prototypes (for GNG and GWR) or the whole of training data-set (for SVM and KNN)  were used with their respective labels to classify the validation data-set. 
As the two latter methods are classic classifier methods, an explanation of SVMs \cite{Boser1992} or KNNs \cite{1053964} is outside the scope of this paper. 
For the GNG and for GWR methods, however maybe some more detail will be given in subsection \ref{class_}, as those methods are topological descriptors of data and not classifiers per se. 
SVM and KNN classifiers were chosen because they were shown~\cite{amancio_systematic_2014} to be the two most accurate methods to classify generic data-sets with higher dimensions. Specially KNN showed an ever increasing performance with increased number of features~\cite[p.6]{amancio_systematic_2014}.
GWR and GNG algorithms were chosen for their high performance on CAD-60 and in HAR~\cite{parisi_human_2014}, offering a promising new way to classify activities. Also, the use of a standard version of the GNG~\cite{konsoulas_unsupervised_2013} enabled us to test our own implementation of the GWR algorithm. 




\subsection{Building training and validation data-set}\label{sec:val}
Choosing the right way to partition validation and training data is very important to give us an accurate estimation of a classifier's performance, as doing an improper division might present  to our classifier an oversimplified learning task and thus, report a non-working classifier as a functional one. 
Conversely, an overly stringent cross-validation strategy, might take a very long time to evaluate (as the classifier needs to be trained multiple times with multiple data partitions) and not present enough data for the learning algorithm to generalise its response, thus making it unable to learn the desired classification task. 

According to Zhang et al.~\cite{act_rgbd}, the de-facto validation scheme is leave one subject out. 
This is perhaps for its simplicity in implementation, but also as it seems to be a robust and sensible way to perform cross-validation as it emulates what would be like to do a real live test, that is, to have a trained classifier model and test it on a complete new subject. 
The limitations of this cross-validation method are that this testing is done with the same instructions on how to perform the actions, same objects, hardware and software implementations, sensor position and background.  


As the data-set has 4 different subjects, this approach implied that we separate the data 4 times, each time excluding one subject from the training set and used the classifier's response to this out-of-sample subject to estimate its accuracy. 
Here the SVM, KNN, GNG and GWR responses calculated 4 times, each time excluding one of the subjects from the set and then averaging the results. 


\subsection{Data representation}
More thorough descriptions~\cite{nite} of the data obtained from the depth sensor should be referenced, but basically this is a set of J points (where J is the number of joints)  with x, y and z coordinates, each representing a landmark on the body in time in a 3D space. A right-handed coordinate system was used with the Y-axis corresponding to height or a vertical displacement, the X-axis corresponds to width or a lateral displacement and the Z-axis corresponding to depth.
We represent thus a particular pose as the concatenation of these J points, such as that for each time frame $k$ we have a pose p represented by the matrix:
\begin{equation}
p(k) =\begin{bmatrix}
j_{1x}(k) & j_{1y}(k) & j_{1z}(k)\\ 
j_{2x}(k) & j_{2y}(k) & j_{2z}(k)\\ 
\dots \\ 
j_{Jx}(k) & j_{Jy}(k) & j_{Jz}(k)
\end{bmatrix} 
\end{equation}
An action sequence represented on discrete time steps $1..K$ could therefore represented as the multidimensional array resulting of the sequential concatenation of the k-th pose matrices. 
To use the pose information with a gas we change the representation of the pose matrix $p(k)$ into a vector size $3J$ and the action sequence is the horizontal concatenation of the all the k-th, $p(k)$ matrices. One may thus understand the the pose vector as a single point in a high dimensional space and an action sequence as a necessarily continuous trajectory in that space.


\subsection{Preconditioning}
Preconditioning of the data before feeding into the learning algorithms was done in 3 different manners taken from current state of the art classifiers for the data-set according to the CAD-60 Cornell's website~\cite{_cornell_????}. 

\subsubsection{Centring and mirroring}
According to Parisi et al., a reasonable preconditioning is to centre the skeleton pose on its hips and concatenate this set with its mirror image across the X-axis, so that the final data-set is twice the size and contains descriptions of joint positions with a sagittal plane symmetry. 
This is done as to account for the lateralisation on performance of actions, that is, actions that were done with a left as opposed to right hand or foot. 

As one might may reason, geometrically mirroring across the X-axis only does what we would hope for in respects to lateralisation (that is, keep the skeleton pose and transform an action performed with the left for one with the right hand) if the skeleton has its mid-sagittal plane passing the origin and its normal vector on X-axis direction. Only the displacement is corrected by the centring procedure, so mirroring will result in a pose that is lateralised but has a different rotation.

\subsubsection{Centring, mirroring and normalizing}
Cippitelli's classifier \cite{cippitelli_human_2016} implements a further step in pre-conditioning the data before it is fed into the learning algorithm, namely a normalisation phase that should account for a more standard representation of actions by people with different sizes. 
This is achieved by proposing that the skeletons should be normalised by the distance between neck and torso.

 Note that from the literature we examined, even though this is the most well thought through preconditioning strategy implemented, it does not account for people with different body proportions. It is known that sitting height ratio, that is ratio of sitting height (trunk length + head length) to stature X 100, varies considerably with normal ageing as well as with different geographic groups \cite{bogin_leg_2010}. In this paper, Bogin cites the work of Eveleth and Tanner that mapped individuals of different ages (from older than 1 year of age to 20 years or more) of different geographic backgrounds and has shown that sitting height ratios of one-year-olds can range from 59 to 64 while this value is in the range of 47-54 for adults over 20 year of age. In the most extreme case, this is a difference can be of around 35\% or more and most likely extends to multiple body parts. 

A special consideration should be made as the only reference we could find about limb lengths was done to demonstrate changes in childhood, but one should expect that such changes continue to occur later in life as body posture changes through senescence and the fact that this system is supposed to be used with older adults could affect relative limb lengths as well as poses and movement speed.



\subsection{Remapping stage}

An unsupervised classifying or clustering algorithm such as neural gas or a growing when required neural gas, remaps the input data into prototypical representations, acting simultaneously as: a data compression mechanism - as we don't need to keep the whole set, but only a fraction of it; a means of improving generalisation, as the mapping procedure is intended to keep only the necessary features we want about the data; as a filter, as small acquisition errors will be disregarded as they will match to the same node. This work implemented the following strategies in regard to remapping, namely:  


\subsubsection{No remapping}
The trivial solution is to do no remapping at all and feed direct data to the learning algorithm. 


\subsubsection{Growing Neural Gas}
We use the implementation of a GNG from MATLAB File Exchange~\cite{konsoulas_unsupervised_2013}.

\subsubsection{Growing When Required Neural Gas}
We use our own implementation of the GWR based on the paper from Parisi et al.~\cite{parisi} and description from Marsland~\cite{marsland}.

\subsection{Labelling}
After the remapping procedure - if one was implemented, the prototypical nodes obtained need to be labelled.
This was done with a KNN with $ k=1 $  using the labels and data from the training data-set to predict the labels of the prototypical skeleton poses or prototypical concatenated pose sequences.


\subsection{Classification stage}\label{class_}

For the GNG and GWR methods we followed the structure from Parisi et al., generating a smaller set from the original training data-set, a set of prototype skeletons (or short prototype skeleton sequences for layers 2 and 3) that are then matched to the input. As in a self-organising map, the neighbourhood relations of these nodes, or in our case, skeleton prototypes, is produced, but not used. The skeleton poses may also be concatenated together in a sliding window manner, and this was done repetitively in the second and third layers, with a window size of 3 samples. This structure was repeated in 2 layers in 2 parallel branches, one for poses and another for velocities, and in an end third layer that concatenates velocities and poses. This was responsible for presenting the best matched combined action sequence. For each compressor layer, the maximum number of prototype skeleton poses or concatenated poses is usually set to 1000, a value obtained from literature and tested experimentally. With an inclusion parameter $a_{T}=1$ and $nodes = \infty$ , the GNG and the GWR would degenerate into a KNN 
 with the number of poses of the size of the training data-set - typically around 30k skeleton poses varying depending on which subject was excluded from the training set. Multiple epochs of the gas construction phase were run to allow the gas to ``set'' into a more consistent state. Additionally, as implemented by Parisi et al., for each epoch, pose examples that didn't reach a threshold level activation of some multiples (set to $\gamma=4$) of the standard deviation of activations were skipped and not added as a new gas node. During classification phase, these prototypes are matched to ones to which they best resemble using a distance function. For this, a simple high dimension Euclidean metric was used, considering our J joints skeleton as a point in a $3J$ higher dimensional space to calculate similarity. The classifier output is then the label of prototype to which our skeleton or skeleton sequence has a smallest possible distance. For a more detailed explanation of such structure, we suggest reading~\cite{parisi,marsland,klein_2016}.

\subsection{Determination of the best running parameters}

Each of the used classifiers was run multiple times to determine best running parameters for each method. 
In using SVMs to create a multi-class classifier, the standard MATLAB implementation of error-correcting output codes (ECOC)\cite{allwein_reducing_2000} procedure was used and SVMs with a linear kernel and radial kernels were attempted, with the former giving the best results.
 In using the KNN classifier, the number $k$ was tested for both centred, mirrored and centred, mirrored and normalized preconditioning with small numbers of K from 1 to 20 and in 20 steps until $k=1000$, however, since these values changed enormously depending on preconditioning used and number of classes of the multi-class classifier, the value of $k = 1$ was used. 
The GWR method, as the GNG presents a chaotic behaviour (the method has the name of gas as the nodes jitter around optimal points), multiple runs were tried with the starting point being the parameters used by Parisi et al., but as his accuracy could not be achieved with our implementation, other parameters were used, with best results with reasonable running time being achieved with $a_{T}=0.995$, $ nodes = 1000$ and $epochs = 10$. 
The remaining parameters were left unchanged, that is learning rates $\epsilon_{b} = 0.2$, $\epsilon_{n} = 0.006$, maximum age threshold $a_{max} = 50$, firing counter $h_{0} =1$ and habituation parameters $\alpha_b=0.95$, $\alpha_n=0.95$ $\tau_b=3.33$. 
GNG was not optimised, as the method its implementation is much slower than all others and was run using parameters for number of nodes and epochs from GWR method ($\lambda=3$,  $\epsilon_{b} = 0.2$, $\epsilon_{n} = 0.006$, $a_{max} = 1$, $d=0.995$), quite similar to the original implementation from Fritzke\cite{fritzke_growing_1995}. 
Varying of the sliding window lengths was not performed for any algorithm. 

\section{Results and Discussion}


The classifiers were run and compared in a ``by scene'' manner, in which the available actions to be classified were not all the 12 different actions, but given a particular scene, say bathroom or bedroom, a smaller set containing between 3 and 4 classes of actions was examined.

The results from each of the 4 methods and the effects of varying preconditioning can be seen on  
table~\ref{tab:results}. This was done to separate the effects of pre-processing stage from the learning algorithms efficacy. For a simpler comparison on which was the best method of all, the overall global  accuracies were used, here defined as the sum of the traces of the confusion matrices of all scenes and all subjects divided by the total number of poses. Appendix~\ref{ap:acc}, describes better the procedure used for this. As the method KNN has shown to be the more accurate one, details of its accuracy by activity type can be seen in table~\ref{tab:winner}.

Regarding the optimal value for $k$ on the KNN classifier: this value seemed to vary considerably. Depending on how many classes there were to be chosen from, its optimal value ranged from around 320 (when all tasks are possible classes) to the optimal value 8 (when there are only 3-4 possible tasks using only centring and mirroring preconditioning) and to the optimal $k = 2$ (when centring, mirroring and normalizing was used). As the difference between the maximum value achieved and the value for $ k=1$ was usually around 0.5\% or less (83.65\% vs. 83.02\% and 82.91\% vs 82.48\% respectively) in those particular cases, we chose to report the outcomes with $ k=1$, which seem to yield a simpler classifier with accuracy very close to the optimal. 

The table~\ref{tab:results} shows clearly that disregarding the translation of the skeleton pose in regard to the origin is advantageous in every method, as well as indicate that the  idea to normalise the skeleton based on neck to torso distance is in most cases helpful - however not in the best performing method only by a small margin (83.02\% against 82.48\%). 
The interesting results when analysed by scene, where clearly in some environments as ``bedroom'' we have a lower than expected precision and recall, were not only occurring in the KNN approach, but in both the gas implementations as well as in the SVM.

  
\begin{table}\caption{Overall global accuracy (in \%) for all actions and all subjects}
\begin{center}
\resizebox{\columnwidth}{!}{
\begin{tabular}{ l c c c c }
 \toprule
  				&  	SVM	 	&	KNN		&	GNG		&	GWR 		\\
 \midrule 
No preconditioning			& 	55.43\%		& 	53.76\%		&	46.44\%	& 	47.27\%	\\   
Centring and mirroring 	&	66.36\% 	&	83.02\% 	& 	75.97\%		& 	75.2\%	\\
Centring, mirroring and normalizing				&	67.27\% 	&	82.48\% 		& 	78.37\%		& 	79\%  	\\
\bottomrule
\end{tabular}\label{tab:results}
}
\end{center}
\end{table}

\begin{table}\caption{Precision and recall of the best-found algorithm (1NN, centred and mirrored skeletons) in the different environments of CAD-60 for all subjects combined.}

\begin{center}
\resizebox{\columnwidth}{!}{
\begin{tabular}{ l c c c }
 \toprule
& & \multicolumn{2}{c}{``New-person''} \\
  Location &  Activity & Precision	&	Recall	\\
 \midrule 
\multirow{4}{.5in}{Bathroom}& Brushing teeth	&$96.46\%	 \pm 4.29\% $ &$ 93.11\%	 \pm 6.84\%	$	\\   
& Rinsing mouth	& $ 87.68\%	 \pm 18.55\% $ & $	100.00\%	 \pm 0.00\% $\\
& Wearing contact lens & $ 99.11\%	 \pm 0.61\% $ & $  88.82\%	 \pm 11.67\% $\\
& Average & $ 94.42\%	 \pm 11.18\% $ & $ 93.98\%	 \pm 8.55\% $\\
\midrule
\multirow{4}{.5in}{Bedroom} & Talking on phone & $ 63.12\%	 \pm 46.84\%	 $ &  $ 60.44\%	 \pm 41.24\%	 $\\
& Drinking water	& $ 55.47\%	 \pm 39.08\%	 $ &	 $ 65.81\%	 \pm 45.57\%	 $\\
& Opening pill container	&   $ 98.83\%	 \pm 2.34\%	 $ & $ 77.85\%	 \pm 39.14\%	 $	\\
& Average & $ 72.47\%	 \pm 37.50\%	 $ & $ 68.03\%	 \pm 38.80\%	 $\\
\midrule
\multirow{5}{.5in}{Kitchen} & Cooking-chopping 	& $ 96.86\%	 \pm 1.94\%	 	$ & $ 75.97\%	 \pm 18.23\% $	\\
& Cooking-stirring						& $ 56.35\%	 \pm 39.66\%	$ & $ 81.43\%	 \pm 29.14\% $	\\
& Drinking water 						& $ 73.41\%	 \pm 48.97\%	$ & $ 74.46\%	 \pm 49.65\% $	\\
& Opening pill container					& $ 98.83\%	 \pm 2.34\%	 	$ & $ 81.06\%	 \pm 32.72\% $	\\
& Average 							& $ 81.36\%	 \pm 33.54\%	$ & $ 78.23\%	 \pm 30.88\% $	\\
\midrule
\multirow{5}{.6in}{Living room} & Talking on phone 	& $ 63.12\%  \pm 46.84\%	$ & $ 60.44\%	 \pm 41.24\% $ 	\\
& Drinking water						& $ 81.28\%	 \pm 17.82\%	$ & $ 79.30\%	 \pm 23.43\% $	\\
& Talking on couch						& $ 100.00\% \pm 0.00\%	$ & $ 99.42\%	 \pm 1.17\%	 $	\\
& Relaxing on couch					&  $ 100.00\% \pm 0.00\%	$ & $ 100.00\%	 \pm 0.00\%	 $	\\
& Average 							& $ 86.10\%	 \pm 27.43\%	$ & $ 84.79\%	 \pm 27.11\%$ 	\\
\midrule
\multirow{5}{.5in}{Office} & Talking on phone 	& $ 63.12\%	 \pm 46.84\%	$ & $ 60.44\%	 \pm 41.24\%$	 \\
& Writing on whiteboard					& $ 89.64\%	 \pm 20.23\%	$ & $ 100.00\%	 \pm 0.00\%	 $	\\
& Drinking water						& $ 82.06\%	 \pm 18.78\%	$ &  $ 76.45\%	 \pm 26.94\% $	\\
& Working on computer					&  $ 100.00\% \pm 0.00\%	$ & $ 100.00\%	 \pm 0.00\%	 $	\\
& Average 							& $ 83.71\%	 \pm 28.02\%	$ & $ 84.22\%	 \pm 28.02\%	 $ \\
\midrule
& Global average 						& $ 94.42\%	 \pm 11.18\%	 $ & $ 81.95\%	 \pm 28.82\%	 $ \\
\bottomrule
\end{tabular}\label{tab:winner}
}
\end{center}

\end{table}

\begin{figure}[!t]
\centering
\includegraphics[width=3.5in]{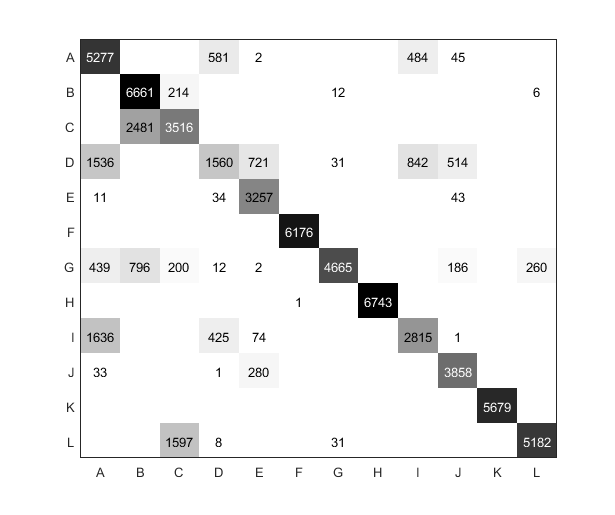}
\caption{Combined confusion matrix of the best classifier found, the 1NN, centred and mirrored skeletons. Letters A through N represent the actions: 'brushing teeth',
'cooking (chopping)',
'cooking (stirring)',
'drinking water',
'opening pill container',	
'random',
'relaxing on couch',
'rinsing mouth with water',
'still',
'talking on couch',
'talking on the phone',
'wearing contact lenses',
'working on computer'	and
'writing on whiteboard'   respectively.}
\label{fig_sim}
\end{figure}
We can also see the combined confusion matrix for all subjects of the best-found classifier in Fig.~\ref{fig_sim}, which was classified not on a by scene basis, but assuming all actions were possible. This is a slight different condition than the one reported on previous tables and yielded a slightly lower global accuracy of 80.36\%.

\subsection{Internal validity}
To assure reproducible results, standard MATLAB functions were used whenever possible; whenever possible data inspection was performed (with plotting of the skeletons and visually assuring that was being implemented was what we were planning to implement), classifiers were run multiple times to account for chaotic behaviour of the gas classifiers. However extensive report of variation of gas parameters and the inherent multi-variable optimization it entails would present a massive and lengthy task, which was not performed.

One should also add that determining the best running parameters for each of the methods was done manually, since classifier implementations were rather slow, making it impossible to use methods for non-linear multivariate optimization. This implies that we cannot guarantee for some of the models which have multiple running parameters (namely the hierarchical ones, i. e., the GNG and GWR) that the same algorithms with a few tweaks are not going to perform much better.
 An additional difficulty with gas-based classifiers is their stochastic nature, in which the final gas obtained may differ a lot based on the initial nodes used, as well as the order in which poses and actions are presented to it, giving varying results. Because of that, they also suffer from possible cherry-picking in the event one may choose to use only the gas ``that learned the task best'', which would inadvertently bypass our cross-validation procedure and not representing a real accuracy.

Another issue that must be addressed is that perhaps using the same architecture to classify all activities is most likely not the best strategy. Different actions have different aspects of them that are important and basing on the assumption that one algorithm will the one that will classify them all in the best way is probably not correct, but a simplification of the problem. A sounder approach would be to optimize a given classifier for each action (along with its running parameters), choose the fittest one for each task and use an additional layer to combine the outputs.  

An extra remark must be made regarding the large variability of the results. The source of this was either one or two subjects having consistently worse results than others (namely subject 2 and 3, one of them being a left handed person), influence that our mirroring procedure did not seem to extinguish. Perhaps a larger data-set would be needed as to allow the algorithm to fill in the gaps of possible variations for a certain performed action or perhaps a better procedure for dealing with handedness should be tried. 
\subsection{External validity}
The results obtained apply to classification of skeleton pose sets of full body motion in an environment that allows for non-occluded or limited self-occlusion of the subject being observed. Moreover, the classification of more fine-grained activities would be a challenge with such setup, considering the inaccuracies of the skeleton poses obtained by the KINECT, as well as a limited capacity of our proposed algorithms to generalise activities. 

The approach used does allow for real time classification, as with even the sliding window algorithms, the window needed is small in comparison to duration of the action, that is 9 frames at the most with a sensor acquiring 30 frames per second allowing us to have a result for current action after 300ms of acquisition. This is an advantage for implementation, as in real use, one will not have the actual beginning or end of an action defined anywhere. In fact, the algorithms run with the structure proposed showed little (if any) advantage on examining a larger number of samples and in fact, the best accuracy observed was that of examining only the current skeleton pose (the first layer used as a classifier would yield better results than the  third combined layer), meaning that the best classifier on the subset we examined can output at most every 33ms the current action. As a matter of fact, the chosen best algorithm, the KNN would output a class for each pose, that is, after 3ms.

As with its recommendation for using with different data-sets not the CAD-60, one might point out the limitation that the skeleton definitions used only 15 joint positions, they are noisy and they would not discriminate fine motor action. Moreover, we do not abstract rotations or take into account persons with different builds or body proportions. One final note before use regards the context of scene, where only a smaller set of actions are considered when classifying a scene, and introducing a classifier with a higher number of actions would probably degrade results quite quickly, taking into account the high variability of the results of both precision and recall among different participants. 

 One should notice that k-folds, the standard cross-validation procedure from MATLAB does not work in our case, as we need to remove samples per subject and classifying it with random removal of samples of each action sequence poses a much simpler and misleading problem, as the samples from each sequence are plenty and they are much too similar to each other in the same action sequence, but considerably different when other actions sequences are considered. This would output a classifier that overfits and does not show this overfitting properties in the cross-validation step. However we cannot claim to have the best possible cross-validation procedure as other types of cross-validation, such as cross-data-set validation as suggested by Zhang et al., were not implemented. And as previously pointed out~\cite{torralba_unbiased_2011}, data-set selection might have an important effect on results (classified the data-set successfully, not the task) and may hide an even greater difficulty for our classifier to generalize. 

\section{Conclusion}
From the analysis we performed there seems to be a law of diminishing returns happening when more complex methods are used to analyse skeleton data. The simple KNN algorithm with just centring the skeletons gives results that are on our implementation the best accuracy overall. The benefits from using methods seem to require some sensitive fine tuning that, with our code, we could not replicate. 

These findings, point in the direction of using simpler methods for this task, perhaps combined with other strategies to limit the classes such as using a state machine that encodes knowledge about the task, or other information, such as the detection of objects that are being grasped or sound (talking on the phone and drinking water would be much more easily discriminated in that manner, as for with pose detection, considerations of where the hand is when the head is modelled as a single point, can be tricky for this approach) and this multi-modal information could be easily integrated in an additional layer to correct inaccuracies of only performing HAR with pose detection. This is our planned future work. 


%

\appendices

\section{Reported average accuracy}\label{ap:acc}
The reported average value of accuracy on all algorithms was computed as the average of all scenes of the average for all subjects of trace of all confusion matrices divided by the total number of poses scanned, that is, for a given method in each scene s for a given participant p, with the confusion matrix C, the average accuracy $A_{method,s,p}$ was calculated as:
\begin{equation}
A_{method,s,p} = tr(C_{method,s,p})/\sum \sum C_{method,s,p}
\end{equation}

The average of all participants was then calculated, letting P be the total number of participants (in our case 4) 
 that is, if $A_{s}$: 
\begin{equation}
A_{method,s} = \sum A_{method,s,p}/P
\end{equation}
Finally, letting S being the total number of scenes, for a given method the average for all scenes was calculated as:
\begin{equation}
A_{method} = \sum A_{method,s}/S
\end{equation}
Which are the overall global accuracy values reported when comparing methods in table~\ref{tab:results}.

\section*{Acknowledgement}

This work was partially supported by CNPq Brazil (scholarship 232590/2014-1).

\ifCLASSOPTIONcaptionsoff
  \newpage
\fi



\bibliographystyle{IEEEtran}
\bibstyle{IEEEtran}
%
\bibliography{final_version}

\begin{thebibliography}{10}
\providecommand{\url}[1]{#1}
\csname url@samestyle\endcsname
\providecommand{\newblock}{\relax}
\providecommand{\bibinfo}[2]{#2}
\providecommand{\BIBentrySTDinterwordspacing}{\spaceskip=0pt\relax}
\providecommand{\BIBentryALTinterwordstretchfactor}{4}
\providecommand{\BIBentryALTinterwordspacing}{\spaceskip=\fontdimen2\font plus
\BIBentryALTinterwordstretchfactor\fontdimen3\font minus
  \fontdimen4\font\relax}
\providecommand{\BIBforeignlanguage}[2]{{%
\expandafter\ifx\csname l@#1\endcsname\relax
\typeout{** WARNING: IEEEtran.bst: No hyphenation pattern has been}%
\typeout{** loaded for the language `#1'. Using the pattern for}%
\typeout{** the default language instead.}%
\else
\language=\csname l@#1\endcsname
\fi
#2}}
\providecommand{\BIBdecl}{\relax}
\BIBdecl

\bibitem{parisi}
\BIBentryALTinterwordspacing
G.~I. Parisi, C.~Weber, and S.~Wermter, ``Self-organizing neural integration of
  pose-motion features for human action recognition,'' \emph{Frontiers in
  Neurorobotics}, vol.~9, Jun. 2015. [Online]. Available:
  \url{http://journal.frontiersin.org/Article/10.3389/fnbot.2015.00003/abstract}
\BIBentrySTDinterwordspacing

\bibitem{chaquet_survey_2013}
\BIBentryALTinterwordspacing
J.~M. Chaquet, E.~J. Carmona, and A.~Fern{\'a}ndez-Caballero, ``A survey of
  video datasets for human action and activity recognition,'' \emph{Computer
  Vision and Image Understanding}, vol. 117, no.~6, pp. 633--659, Jun. 2013.
  [Online]. Available:
  \url{http://www.sciencedirect.com/science/article/pii/S1077314213000295}
\BIBentrySTDinterwordspacing

\bibitem{act_rgbd}
\BIBentryALTinterwordspacing
J.~Zhang, W.~Li, P.~O. Ogunbona, P.~Wang, and C.~Tang, ``{RGB}-{D}-based
  {Action} {Recognition} {Datasets}: {A} {Survey},'' \emph{arXiv preprint
  arXiv:1601.05511}, 2016. [Online]. Available:
  \url{http://arxiv.org/abs/1601.05511}
\BIBentrySTDinterwordspacing

\bibitem{gupta_human_2013}
\BIBentryALTinterwordspacing
R.~Gupta, A.~Y.-S. Chia, and D.~Rajan, ``Human activities recognition using
  depth images,'' in \emph{Proceedings of the 21st {ACM} international
  conference on Multimedia}.\hskip 1em plus 0.5em minus 0.4em\relax {ACM}, pp.
  283--292. [Online]. Available:
  \url{http://dl.acm.org/citation.cfm?id=2502099}
\BIBentrySTDinterwordspacing

\bibitem{shan_3d_2014}
\BIBentryALTinterwordspacing
J.~Shan and S.~Akella, ``3d human action segmentation and recognition using
  pose kinetic energy,'' in \emph{Advanced Robotics and its Social Impacts
  ({ARSO}), 2014 {IEEE} Workshop on}.\hskip 1em plus 0.5em minus 0.4em\relax
  {IEEE}, pp. 69--75. [Online]. Available:
  \url{http://ieeexplore.ieee.org/abstract/document/7020983/}
\BIBentrySTDinterwordspacing

\bibitem{faria_probabilistic_2014}
D.~R. Faria, C.~Premebida, and U.~Nunes, ``A probabilistic approach for human
  everyday activities recognition using body motion from {RGB}-d images,'' in
  \emph{The 23rd {IEEE} International Symposium on Robot and Human Interactive
  Communication}, pp. 732--737.

\bibitem{cippitelli_human_2016}
\BIBentryALTinterwordspacing
E.~Cippitelli, S.~Gasparrini, E.~Gambi, and S.~Spinsante, ``A human activity
  recognition system using skeleton data from {RGBD} sensors,'' vol. 2016, pp.
  1--14. [Online]. Available:
  \url{http://www.hindawi.com/journals/cin/2016/4351435/}
\BIBentrySTDinterwordspacing

\bibitem{gines_hidalgo_zhe_cao_tomas_simon_shih-en_wei_hanbyul_joo_yaser_sheikh_openpose:_????}
\BIBentryALTinterwordspacing
``{OpenPose}: Real-time multi-person keypoint detection library for body, face,
  and hands estimation.'' [Online]. Available:
  \url{https://github.com/CMU-Perceptual-Computing-Lab/openpose}
\BIBentrySTDinterwordspacing

\bibitem{berry_falls:_2008}
\BIBentryALTinterwordspacing
S.~D. Berry and R.~Miller, ``Falls: Epidemiology, pathophysiology, and
  relationship to fracture,'' vol.~6, no.~4, pp. 149--154. [Online]. Available:
  \url{https://www.ncbi.nlm.nih.gov/pmc/articles/PMC2793090/}
\BIBentrySTDinterwordspacing

\bibitem{masud2001epidemiology}
T.~Masud and R.~O. Morris, ``Epidemiology of falls,'' \emph{Age and ageing},
  vol.~30, no. suppl\_4, pp. 3--7, 2001.

\bibitem{gaenslen_early_2010}
\BIBentryALTinterwordspacing
A.~Gaenslen and D.~Berg, ``Early diagnosis of parkinson’s disease,'' in
  \emph{International Review of Neurobiology}, ser. Transcranial Sonography in
  Movement Disorders, D.~Berg and U.~Walter, Eds.\hskip 1em plus 0.5em minus
  0.4em\relax Academic Press, vol.~90, pp. 81--92, {DOI}:
  10.1016/S0074-7742(10)90006-8. [Online]. Available:
  \url{http://www.sciencedirect.com/science/article/pii/S0074774210900068}
\BIBentrySTDinterwordspacing

\bibitem{klucken_unbiased_2013}
\BIBentryALTinterwordspacing
J.~Klucken, J.~Barth, P.~Kugler, J.~Schlachetzki, T.~Henze, F.~Marxreiter,
  Z.~Kohl, R.~Steidl, J.~Hornegger, B.~Eskofier, and J.~Winkler, ``Unbiased and
  mobile gait analysis detects motor impairment in parkinson's disease,''
  vol.~8, no.~2, p. e56956. [Online]. Available:
  \url{http://journals.plos.org/plosone/article?id=10.1371/journal.pone.0056956}
\BIBentrySTDinterwordspacing

\bibitem{cecil}
\BIBentryALTinterwordspacing
L.~G. MD and A.~I.~S. MD, \emph{Goldman-Cecil Medicine, 2-Volume Set, 25e
  (Cecil Textbook of Medicine)}.\hskip 1em plus 0.5em minus 0.4em\relax
  Elsevier, 2015. [Online]. Available:
  \url{https://www.amazon.com/Goldman-Cecil-Medicine-Set-Cecil-Textbook/dp/1455750174?SubscriptionId=0JYN1NVW651KCA56C102&tag=techkie-20&linkCode=xm2&camp=2025&creative=165953&creativeASIN=1455750174}
\BIBentrySTDinterwordspacing

\bibitem{di_lenarda_future_2017}
\BIBentryALTinterwordspacing
A.~Di~Lenarda, G.~Casolo, M.~M. Gulizia, N.~Aspromonte, S.~Scalvini,
  A.~Mortara, G.~Alunni, R.~P. Ricci, R.~Mantovan, G.~Russo, G.~F. Gensini, and
  F.~Romeo, ``The future of telemedicine for the management of heart failure
  patients: a consensus document of the italian association of hospital
  cardiologists (a.n.m.c.o), the italian society of cardiology (s.i.c.) and the
  italian society for telemedicine and {eHealth} (digital s.i.t.),'' vol.~19,
  pp. D113--D129. [Online]. Available:
  \url{https://www.ncbi.nlm.nih.gov/pmc/articles/PMC5520762/}
\BIBentrySTDinterwordspacing

\bibitem{demaerschalk_stroke_2009}
\BIBentryALTinterwordspacing
B.~M. Demaerschalk, M.~L. Miley, T.-E.~J. Kiernan, B.~J. Bobrow, D.~A. Corday,
  K.~E. Wellik, M.~I. Aguilar, T.~J. Ingall, D.~W. Dodick, K.~Brazdys, T.~C.
  Koch, M.~P. Ward, and P.~C. Richemont, ``Stroke telemedicine,'' vol.~84,
  no.~1, pp. 53--64. [Online]. Available:
  \url{https://www.ncbi.nlm.nih.gov/pmc/articles/PMC2664571/}
\BIBentrySTDinterwordspacing

\bibitem{cooper_respiratory_2009}
\BIBentryALTinterwordspacing
C.~B. Cooper, ``Respiratory applications of telemedicine,'' vol.~64, no.~3, pp.
  189--191. [Online]. Available: \url{http://thorax.bmj.com/content/64/3/189}
\BIBentrySTDinterwordspacing

\bibitem{flodgren_interactive_2015}
\BIBentryALTinterwordspacing
G.~Flodgren, A.~Rachas, A.~J. Farmer, M.~Inzitari, and S.~Shepperd,
  ``Interactive telemedicine: effects on professional practice and health care
  outcomes,'' in \emph{Cochrane Database of Systematic Reviews}.\hskip 1em plus
  0.5em minus 0.4em\relax John Wiley \& Sons, Ltd. [Online]. Available:
  \url{http://onlinelibrary.wiley.com/doi/10.1002/14651858.CD002098.pub2/abstract}
\BIBentrySTDinterwordspacing

\bibitem{cad60}
\BIBentryALTinterwordspacing
J.~Sung, C.~Ponce, B.~Selman, and A.~Saxena, ``Unstructured human activity
  detection from rgbd images,'' in \emph{Robotics and {Automation} ({ICRA}),
  2012 {IEEE} {International} {Conference} on}.\hskip 1em plus 0.5em minus
  0.4em\relax IEEE, 2012, pp. 842--849. [Online]. Available:
  \url{http://ieeexplore.ieee.org/abstract/document/6224591/}
\BIBentrySTDinterwordspacing

\bibitem{_cornell_????}
\BIBentryALTinterwordspacing
Cornell activity dataset ({CAD}-60) results. available at:
  \url{http://pr.cs.cornell.edu/humanactivities/results.php} accessed in
  28\textsuperscript{th} july 2017. [Online]. Available:
  \url{http://pr.cs.cornell.edu/humanactivities/results.php}
\BIBentrySTDinterwordspacing

\bibitem{alessandro_manzi_human_2017}
\BIBentryALTinterwordspacing
{Alessandro Manzi}, {Paolo Dario}, and {Filippo Cavallo}, ``A human activity
  recognition system based on dynamic clustering of skeleton data,'' vol.~17,
  no.~5, p. 1100. [Online]. Available:
  \url{http://www.mdpi.com/1424-8220/17/5/1100}
\BIBentrySTDinterwordspacing

\bibitem{klein_2016}
\BIBentryALTinterwordspacing
F.~B. Klein, K.~Štěpánová, and A.~Cangelosi, ``Implementation of a modular
  growing when required neural gas architecture for recognition of falls,'' in
  \emph{Neural Information Processing}, ser. Lecture Notes in Computer Science,
  A.~Hirose, S.~Ozawa, K.~Doya, K.~Ikeda, M.~Lee, and D.~Liu, Eds.\hskip 1em
  plus 0.5em minus 0.4em\relax Springer International Publishing, 2016, no.
  9947, pp. 526--534, {DOI}: 10.1007/978-3-319-46687-3\_58. [Online].
  Available:
  \url{http://link.springer.com/chapter/10.1007/978-3-319-46687-3\_58}
\BIBentrySTDinterwordspacing

\bibitem{noauthor_KINECT_nodate}
\BIBentryALTinterwordspacing
``Kinect for {Windows} {Sensor} {Components} and {Specifications}.'' [Online].
  Available: \url{https://msdn.microsoft.com/en-us/library/jj131033.aspx}
\BIBentrySTDinterwordspacing

\bibitem{Boser1992}
\BIBentryALTinterwordspacing
B.~E. Boser, I.~M. Guyon, and V.~N. Vapnik, ``A training algorithm for optimal
  margin classifiers,'' in \emph{Proceedings of the fifth annual workshop on
  Computational learning theory}.\hskip 1em plus 0.5em minus 0.4em\relax {ACM}
  Press, 1992. [Online]. Available: \url{https://doi.org/10.1145/130385.130401}
\BIBentrySTDinterwordspacing

\bibitem{1053964}
T.~Cover and P.~Hart, ``Nearest neighbor pattern classification,'' \emph{IEEE
  Transactions on Information Theory}, vol.~13, no.~1, pp. 21--27, January
  1967.

\bibitem{amancio_systematic_2014}
\BIBentryALTinterwordspacing
D.~R. Amancio, C.~H. Comin, D.~Casanova, G.~Travieso, O.~M. Bruno, F.~A.
  Rodrigues, and L.~d.~F. Costa, ``A systematic comparison of supervised
  classifiers,'' vol.~9, no.~4, p. e94137. [Online]. Available:
  \url{http://journals.plos.org/plosone/article?id=10.1371/journal.pone.0094137}
\BIBentrySTDinterwordspacing

\bibitem{parisi_human_2014}
\BIBentryALTinterwordspacing
G.~I. Parisi, C.~Weber, and S.~Wermter, ``Human action recognition with
  hierarchical growing neural gas learning,'' in \emph{Artificial Neural
  Networks and Machine Learning–{ICANN} 2014}.\hskip 1em plus 0.5em minus
  0.4em\relax Springer, pp. 89--96. [Online]. Available:
  \url{http://link.springer.com/chapter/10.1007/978-3-319-11179-7_12}
\BIBentrySTDinterwordspacing

\bibitem{konsoulas_unsupervised_2013}
\BIBentryALTinterwordspacing
I.~Konsoulas, ``Unsupervised {Learning} with {Growing} {Neural} {Gas} ({GNG})
  {Neural} {Network} - {File} {Exchange} - {MATLAB} {Central},'' Sep. 2013.
  [Online]. Available:
  \url{http://www.mathworks.com/matlabcentral/fileexchange/43665-unsupervised-learning-with-growing-neural-gas--gng--neural-network}
\BIBentrySTDinterwordspacing

\bibitem{nite}
\BIBentryALTinterwordspacing
Prime sensor™ {NITE} 1.3 framework programmer's guide - {NITE}.pdf. [Online].
  Available: \url{http://pr.cs.cornell.edu/humanactivities/data/NITE.pdf}
\BIBentrySTDinterwordspacing

\bibitem{bogin_leg_2010}
\BIBentryALTinterwordspacing
B.~Bogin and M.~I. Varela-Silva, ``Leg length, body proportion, and health: A
  review with a note on beauty,'' vol.~7, no.~3, pp. 1047--1075. [Online].
  Available: \url{https://www.ncbi.nlm.nih.gov/pmc/articles/PMC2872302/}
\BIBentrySTDinterwordspacing

\bibitem{marsland}
\BIBentryALTinterwordspacing
S.~Marsland, J.~Shapiro, and U.~Nehmzow, ``A self-organising network that grows
  when required,'' \emph{Neural Networks}, vol.~15, no.~8, pp. 1041--1058,
  2002. [Online]. Available:
  \url{http://www.sciencedirect.com/science/article/pii/S0893608002000783}
\BIBentrySTDinterwordspacing

\bibitem{allwein_reducing_2000}
E.~L. Allwein, R.~E. Schapire, and Y.~Singer, ``Reducing multiclass to binary:
  A unifying approach for margin classifiers,'' vol.~1, pp. 113--141.

\bibitem{fritzke_growing_1995}
\BIBentryALTinterwordspacing
B.~Fritzke, ``A growing neural gas network learns topologies,'' vol.~7, pp.
  625--632. [Online]. Available:
  \url{https://www.cs.swarthmore.edu/~meeden/DevelopmentalRobotics/fritzke95.ps}
\BIBentrySTDinterwordspacing

\bibitem{torralba_unbiased_2011}
A.~Torralba and A.~A. Efros, ``Unbiased look at dataset bias,'' in \emph{{CVPR}
  2011}, pp. 1521--1528.

\end{thebibliography}

\end{document}